\PassOptionsToPackage{usenames,dvipsnames,table}{xcolor}

\documentclass[11pt,a4paper]{article}
\usepackage[]{acl2018}
\aclfinaltrue  % for camera-ready

\usepackage{times}
\usepackage{latexsym}
\usepackage{graphicx}
\usepackage{amsmath}% http://ctan.org/pkg/amsmath
\usepackage{balance}  % for  \balance command ON LAST PAGE  (only there!)
\usepackage{mdwlist}
\usepackage{graphics}
\usepackage{color}
\usepackage{rotating}
\usepackage{booktabs}
\usepackage{epsfig}
\usepackage{alltt}
\usepackage{moreverb}
\usepackage{fancyvrb}
\usepackage{enumerate}
\usepackage{colortbl}           % for color table cells
\usepackage{xspace} 
\usepackage{relsize}
\usepackage{array}
\usepackage{txfonts}
\usepackage{amsfonts}
\usepackage{amssymb}
\usepackage[toc,page]{appendix}
\usepackage[labelfont=bf]{caption}
\usepackage{tabularx}% http://ctan.org/pkg/tabularx
\usepackage{caption}
\usepackage{rotating}
\usepackage{latexsym}
\usepackage{multirow}
\usepackage{float}
\usepackage{array,etoolbox}
\usepackage{enumitem}
\usepackage[caption=false]{subfig}
\usepackage{arydshln}
\usepackage{setspace}
\usepackage{float}
\usepackage{bm}
\usepackage{xcolor}
\usepackage[toc,page]{appendix}
\usepackage{soul}
\usepackage[draft]{pgf}
\usepackage{centernot}
\usepackage{makecell}
\usepackage{flexisym} % for \textprime
\usepackage{multirow}

\usepackage{mathtools}

\usepackage{algorithm}
\usepackage[noend]{algpseudocode}

\definecolor{vlgray}{gray}{0.95}

\usepackage{array}
\newcolumntype{L}[1]{>{\raggedright\let\newline\\\arraybackslash\hspace{0pt}}m{#1}}
\newcolumntype{C}[1]{>{\centering\let\newline\\\arraybackslash\hspace{0pt}}m{#1}}
\newcolumntype{R}[1]{>{\raggedleft\let\newline\\\arraybackslash\hspace{0pt}}m{#1}}

\usepackage{lscape}
\newcolumntype{b}{>{\hsize=2.3\hsize}X}
\newcolumntype{s}{>{\hsize=.45\hsize}X}
\newcolumntype{m}{>{\hsize=.9\hsize}X}

\algrenewcommand\algorithmicindent{0.9em}%
\algdef{SE}[SUBALG]{Indent}{EndIndent}{}{\algorithmicend\ }%
\algtext*{Indent}
\algtext*{EndIndent}

\usepackage{etoolbox}
\newtoggle{draft}
\togglefalse{draft}
\iftoggle{draft}{
\newcommand{\tushark}[1]{\textcolor{BurntOrange}{[#1 \textsc{--tushar}]}}
\newcommand{\ashish}[1]{\textcolor{OliveGreen}{[#1 \textsc{--ashish}]}}
\newcommand{\dk}[1]{\textcolor{Maroon}{[#1 \textsc{--dk}]}}
\newcommand{\ed}[1]{\textcolor{Blue}{[#1 \textsc{--ed}]}}
\newcommand{\fillthis}[1]{\textcolor{Red}{[#1]}}
\newcommand{\added}[1]{\textcolor{Blue}{#1}}
}{
\newcommand{\tushark}[1]{}
\newcommand{\ashish}[1]{}
\newcommand{\dk}[1]{}
\newcommand{\ed}[1]{}
\newcommand{\fillthis}[1]{}
\newcommand{\added}[1]{#1}
}

\newcommand{\namecite}[1]{\citeauthor{#1}~\shortcite{#1}}

\newcommand{\method}{\textsc{AdvEntuRe}\xspace}
\newcommand{\disc}{\ensuremath{\mathbb{D}}\xspace}
\newcommand{\discretro}{\ensuremath{\mathbb{D}_{\mathrm{retro}}}\xspace}
\newcommand{\gen}{\ensuremath{\mathbb{G}}\xspace}
\newcommand{\gens}{\ensuremath{\mathbb{G}^{\mathrm{s2s}}}\xspace}
\newcommand{\genr}{\ensuremath{\mathbb{G}^{\mathrm{rule}}}\xspace}
\newcommand{\genk}{\ensuremath{\mathbb{G}^{\mathrm{KB}}}\xspace}
\newcommand{\genh}{\ensuremath{\mathbb{G}^{\mathrm{H}}}\xspace}
\newcommand{\orig}{X\xspace}
\newcommand{\synth}{Z\xspace}
\newcommand{\classes}{\ensuremath{\mathcal{C}}\xspace}
\newcommand{\rules}{\ensuremath{\mathcal{R}}\xspace}
\newcommand{\entails}{\textit{entails}\xspace}
\newcommand{\neutral}{\textit{neutral}\xspace}
\newcommand{\contradicts}{\textit{contradicts}\xspace}

\newcommand{\esym}{\ensuremath{\sqsubseteq}\xspace}
\newcommand{\nsym}{\ensuremath{\#}\xspace}
\newcommand{\csym}{\ensuremath{\curlywedge}\xspace}
\newcommand{\negasnli}{nega-SNLI\xspace}

\newcommand\T{\rule{0pt}{2.2ex}}       % Top strut
\newcommand\Tlarge{\rule{0pt}{3ex}}    % Top strut, large
\newcommand\B{\rule[-1.2ex]{0pt}{0pt}} % Bottom strut
\newcommand\R{\rule{0pt}{2.0ex}}       % extra height

\newcommand{\ignore}[1]{}   % ignore text within \ignore{...}
\title{\textsc{AdvEntuRe}: Adversarial Training for Textual Entailment\\ with Knowledge-Guided Examples}

\author{
\makecell{Dongyeop Kang$^1$\quad Tushar Khot$^2$\quad Ashish Sabharwal$^2$\quad Eduard Hovy$^1$} \\
$^1$School of Computer Science, Carnegie Mellon University, Pittsburgh, PA, USA \\
$^2$Allen Institute for Artificial Intelligence, Seattle, WA, USA\\
{\tt $\{$dongyeok,hovy$\}$@cs.cmu.edu}\quad {\tt $\{$tushark,ashishs$\}$@allenai.org}}

\DeclareMathOperator*{\argmin}{\arg\!\min}
\DeclareMathOperator*{\argmax}{\arg\!\max}

\begin{document}

\maketitle

\begin{abstract}
We consider the problem of learning textual entailment models with limited supervision (5K-10K training examples), and present two complementary approaches for it.  First, we propose knowledge-guided adversarial example generators for incorporating large lexical resources in entailment models via only a handful of rule templates.  Second, to make the entailment model---a discriminator---more robust, we propose the first GAN-style approach for training it using a natural language example generator that iteratively adjusts based on the discriminator's performance.  We demonstrate effectiveness using two entailment datasets, where the proposed methods increase accuracy by 4.7\% on SciTail and by 2.8\% on a 1\% training sub-sample of SNLI. Notably, even a single hand-written rule, \textsc{negate}, improves the accuracy on the negation examples in SNLI by 6.1\%.
\end{abstract}

%%%%%%%%%%%%%%%%%%%%%%%%%%%%%%%%%%%%%%%%%%%%%%%%%%%%%%%%%%%%%%%%%%%%%%%%%%%%%%%%%%%%%%%%%%%%%%%%
\section{Introduction}
%%%%%%%%%%%%%%%%%%%%%%%%%%%%%%%%%%%%%%%%%%%%%%%%%%%%%%%%%%%%%%%%%%%%%%%%%%%%%%%%%%%%%%%%%%%%%%%%

The impressive success of machine learning models on large natural language datasets often does not carry over to moderate training data regimes, where models often struggle with infrequently observed patterns and simple adversarial variations.
A prominent example of this phenomenon is \emph{textual entailment}, the fundamental task of deciding whether a \emph{premise} text entails ($\vDash$) a \emph{hypothesis} text.
On certain datasets, recent deep learning entailment systems~\cite{parikh2016decomposable,wang2017bilateral,gong2017natural} have achieved close to human level performance. Nevertheless, the problem is far from solved, as evidenced by how easy it is to generate minor adversarial examples that break even the best systems. 
As Table~\ref{tab:example} illustrates, a state-of-the-art neural system for this task, namely the Decomposable Attention Model~\cite{parikh2016decomposable}, fails when faced with simple linguistic phenomena such as negation, or a re-ordering of words. This is not unique to a particular model or task. Minor adversarial examples have also been found to easily break neural systems on other linguistic tasks such as reading comprehension~\cite{jia2017adversarial}.

\begin{table}[t] %{\linewidth}
%\vspace{0mm}
\centering
\renewcommand{\arraystretch}{1.0}
\caption{\label{tab:example}Failure examples from the SNLI dataset: negation (Top) and re-ordering (Bottom). \textbf{P} is premise, \textbf{H} is hypothesis, and \textbf{S} is prediction made by an entailment system~\cite{parikh2016decomposable}. }
\begin{tabular}{>{\arraybackslash}l}
\Xhline{2\arrayrulewidth}
\hline
\textbf{P}: The dog did \textcolor{red}{not} eat all of the chickens.\\
\textbf{H}: The dog ate all of the chickens. \\
\textbf{S}: entails (score $56.5\%$) \\
\hline
\textbf{P}: \textcolor{red}{The red box} is in the blue box.\\
\textbf{H}: The blue box is in \textcolor{red}{the red box}.\\
\textbf{S}: entails (score $92.1\%$)\\
\hline 
\Xhline{2\arrayrulewidth}
\end{tabular}
%\vspace{-3mm}
\end{table}

A key contributor to this brittleness is the use of specific datasets such as SNLI~\cite{bowman2015large} and SQuAD~\cite{rajpurkar2016squad} to drive model development. While large and challenging, \emph{these datasets also tend to be homogeneous}. E.g., SNLI was created by asking crowd-source workers to generate entailing sentences, which then tend to have limited linguistic variations and annotation artifacts~\cite{annotationArtifacts}. Consequently, models overfit to sufficiently repetitive patterns---and sometimes idiosyncrasies---in the datasets they are trained on. They fail to cover long-tail and rare patterns in the training distribution, or linguistic phenomena such as negation that would be obvious to a layperson.

To address this challenge, we propose to \emph{train textual entailment models more robustly using adversarial examples} generated in two ways: (a) by incorporating knowledge from large linguistic resources, and (b) using a sequence-to-sequence neural model in a GAN-style framework.

The motivation stems from the following observation.
While deep-learning based textual entailment models lead the pack, they generally do not incorporate intuitive rules such as negation, and ignore large-scale linguistic resources such as PPDB~\cite{ganitkevitch2013ppdb} and WordNet~\cite{miller1995wordnet}. These resources could help them generalize beyond specific words observed during training. For instance, while the SNLI dataset contains the pattern \textit{two men $\vDash$ people}, it does not contain the analogous pattern \textit{two dogs $\vDash$ animals} found easily in WordNet.

Effectively integrating simple rules or linguistic resources in a deep learning model, however, is challenging. Doing so directly by substantially adapting the model architecture~\cite{sha2016recognizing,chen2018natural} can be cumbersome and limiting. Incorporating such knowledge indirectly via modified word embeddings~\cite{faruqui2014retrofitting,mrkvsic2016counter}, as we show, can have little positive impact and can even be detrimental.

Our proposed method, which is task-specific but model-independent, is inspired by data-augmentation techniques. We generate new training examples by applying knowledge-guided rules, via only a handful of rule templates, to the original training examples. Simultaneously, we also use a sequence-to-sequence or seq2seq model for each entailment class to generate new hypotheses from a given premise, adaptively creating new adversarial examples. These can be used with any entailment model without constraining model architecture.

We also introduce
the first approach to train a robust entailment model using a Generative Adversarial Network or GAN~\cite{goodfellow2014generative} style framework. We iteratively improve both the entailment system (the \emph{discriminator}) and the differentiable part of the data-augmenter (specifically the neural \emph{generator}), by training the generator based on the discriminator's performance on the generated examples.
%\tushark{Not really, right ? Maybe rephrase to ``training the generators based on the \emph{easiness} of the generated examples''}\dk{Yes. Corrected.} 
%after each training epoch.
Importantly, unlike the typical use of GANs to create a strong generator, we use it as a mechanism to create a strong and robust discriminator.

Our new entailment system, called \method, demonstrates that in the moderate data regime, adversarial iterative data-augmentation via only a handful of linguistic rule templates can be surprisingly powerful. Specifically, we observe 4.7\% accuracy improvement on the challenging SciTail dataset~\cite{khot2018scitail} and a 2.8\% improvement on 10K-50K training subsets of SNLI. 
An evaluation of our algorithm on the negation examples in the test set of SNLI reveals a 6.1\% improvement from just a single rule.

%%%%%%%%%%%%%%%%%%%%%%%%%%%%%%%%%%%%%%%%%%%%%%%%
\section{Related Work}
%%%%%%%%%%%%%%%%%%%%%%%%%%%%%%%%%%%%%%%%%%%%%%%%
% \dk{The intro has too many related works. Probably move some of related works into Related Work section.}
% This work can be seen as a type of neural-symbolic learning~\cite{besold2017neural} which combines neural method and symbolic method together to take advantages of both schools of learning.
% \citet{liang2016neural} combine neural machines with symbolic supervision for the semantic parsing task. 
% \citet{kang2017detecting} train a sequence-to-sequence neural network on causal/effectual pairs of symbolic knowledge.
% Unlike these efforts, our method (1) injects explicit knowledge or hand written rules and (2) trains an entailment discriminator with the help of knowledge guided generators in an end-to-end fashion. %\ashish{add some distinguishing aspect}
% There are many ways of constraining logical rules: embeddings~\cite{faruqui2014retrofitting}, learning~\cite{hu2016harnessing} and so on.
% \tushark{I am not sure about even having this para about general neural+symbolic integration which is not even the focus of this paper. Maybe talk about data-augmentation instead. Specifically the SQUAD adversarial examples and the new ACL paper on breaking NLI.}
% \dk{Agreed. I merged some of them in the next paragraph and make a new paragraph below regarding the adversarial aspect.
%}

\added{
Adversarial example generation has recently received much attention in NLP. For example,
\citet{jia2017adversarial} generate adversarial examples using manually defined templates for the SQuAD reading comprehension task.
\citet{glockner_acl18} create an adversarial dataset from SNLI by using WordNet knowledge. Automatic methods~\cite{Iyyer2018AdversarialEG} have also been proposed to generate adversarial examples through paraphrasing.
These works reveal how neural network systems trained on a large corpus can easily break when faced with carefully designed unseen adversarial patterns at test time.
Our motivation is different. We use adversarial examples at training time, in a data augmentation setting, to train a more robust entailment discriminator. The generator uses explicit knowledge or hand written rules, and is trained in a end-to-end fashion along with the discriminator. % with the help of knowledge guided generators in an end-to-end fashion.
}

Incorporating external rules or linguistic resources in a deep learning model generally requires substantially adapting the model architecture~\cite{sha2016recognizing,liang2016neural,kang2017detecting}. This is a model-dependent approach, which can be cumbersome and constraining. Similarly non-neural textual entailment models have been developed that incorporate knowledge bases. However, these also require model-specific engineering~\cite{raina2005robust,haghighi2005robust,silva2018recognizing}.

An alternative is the model- and task-independent route of incorporating linguistic resources via word embeddings that are \emph{retro-fitted}~\cite{faruqui2014retrofitting} or \emph{counter-fitted}~\cite{mrkvsic2016counter} to such resources. We demonstrate, however, that this has little positive impact in our setting and can even be detrimental. Further, it is unclear how to incorporate knowledge sources into advanced representations such as contextual embeddings~\cite{mccann2017learned,Peters2018ELMO}. \added{We thus focus on a task-specific but model-independent approach.}

Logical rules have also been defined to label existing examples based on external resources~\cite{hu2016harnessing}. Our focus here is on generating \emph{new} training examples.
%However, it is not clear how these resources can be used to label examples in the entailment datasets.

Our use of the GAN framework to create a better discriminator is related to CatGANs~\cite{wang2017catgan} and TripleGANs~\cite{chongxuan2017triple} where the discriminator is trained to classify the original training image classes as well as a new `fake' image class. We, on the other hand, generate examples belonging to the same classes as the training examples. Further, unlike the earlier focus on the vision domain, this is the first approach to train a discriminator using GANs for a natural language task with discrete outputs.
%(unlike previous approaches in the vision domain). 
%This is related to the use of ``feature matching'' GANs for semi-supervised learning~\cite{salimans2016improved,dai2017good} but differs in generating instances from two different types of generators: (1) seq2seq generators and (2) rule-based generators. \ashish{That difference isn't applicable now. Instead, could highlight that previous work was mostly in the Vision domain (was it?)}

% \dk{Tushar's comments on harnessing paper.
% Looking at the harnessing deep networks paper, I think the key difference is that they are using the rules to produce new labels for real examples whereas we are using the rules to produce new examples from real examples. I don’t think we can use our rules to predict the label of an example reliably, so there doesn’t seem to be a way to compare against that paper.
% However, their Equation (2) is very close to what we are doing. Their loss compares the model predictions to the true labels and rule-based network labels for the  true examples. Our model compares the model predictions to the labels of the true examples and labels for the rule-based examples.}

%%%%%%%%%%%%%%%%%%%%%%%%%%%%%%%%%%%%%%%%%%%%%%%%%%%%%%%%%%%%%%%%%%%%%%%%%%%%%%%%%%%%%%%%%%%%%%%%
\section{Adversarial Example Generation}
%%%%%%%%%%%%%%%%%%%%%%%%%%%%%%%%%%%%%%%%%%%%%%%%%%%%%%%%%%%%%%%%%%%%%%%%%%%%%%%%%%%%%%%%%%%%%%%%

% \tushark{High-level story line:\\
% - Generate Adversarial examples using R. \\
% - Three instances of R: \\
%  (a) Knowledge Guided (Replace) \\
%  (b) Hand-authored (Modify) \\
%  (c) Neural (Seq2Seq) \\
%  (?) Second-order examples\\
% - Use GAN to iteratively train D and G }

We present three different techniques to create adversarial examples for textual entailment. Specifically, we show how external knowledge resources, hand-authored rules, and neural language generation models can be used to generate such examples. Before describing these generators in detail, we introduce the notation used henceforth.

We use lower-case letters for single instances (e.g., $x, p, h$), upper-case letters for sets of instances (e.g., $X, P, H$), blackboard bold for models (e.g., $\mathbb{D}$), and calligraphic symbols for discrete spaces of possible values (e.g., class labels $\mathcal{C}$). For the textual entailment task, we assume each example is represented as a triple ($p$, $h$, $c$), where $p$ is a premise (a natural language sentence), $h$ is a hypothesis, and $c$ is an entailment label: (a) \entails (\esym) if $h$ is true whenever $p$ is true; (b) \contradicts (\csym) if $h$ is false whenever $p$ is true; or (c) \neutral (\nsym) if the truth value of $h$ cannot be concluded from $p$ being true.\footnote{The symbols are based on Natural Logic~\cite{lakoff1970linguistics} and use the notation of \namecite{MacCartney2012NaturalLA}.}

We will introduce various example generators in the rest of this section. Each such generator, $\gen_\rho$, is defined by a partial function $f_\rho$ and a label $g_\rho$. If a sentence $s$ has a certain property required by $f_\rho$ (e.g., contains a particular string), $f_\rho$ transforms it into another sentence $s'$ and $g_\rho$ provides an entailment label from $s$ to $s'$. Applied to a sentence $s$, $\gen_\rho$ thus either ``fails'' (if the pre-requisite isn't met) or generates a new entailment example triple, $\left(s, f_\rho(s), g_\rho\right)$. For instance, consider the generator for $\rho$:=hypernym(car, vehicle) with the (partial) transformation function $f_\rho$:=``Replace \textit{car} with \textit{vehicle}'' and the label $g_\rho$:=\entails. $f_\rho$ would fail when applied to a sentence not containing the word ``car''. Applying $f_\rho$ to the sentence s=``\textit{A man is driving the car}'' would generate s'=``\textit{A man is driving the vehicle}'', creating the example $(s, s', \entails)$.

The seven generators we use for experimentation are summarized in Table~\ref{tab:func} and discussed in more detail subsequently. While these particular generators are simplistic and one can easily imagine more advanced ones, we show that training using adversarial examples created using even these simple generators leads to substantial accuracy improvement on two datasets.

\begin{table}[t]
\centering
\fontsize{10.5}{11}\selectfont
\begin{tabular}{@{}c|c|c|c@{}}
\T \textbf{Source} & \bm{$\rho$} & \bm{$f_\rho$}(s) & \bm{$g_\rho$} \\
% \hline
% \multicolumn{4}{l}{}\\
% \hline
\rowcolor[gray]{.8}\multicolumn{4}{c}{\Tlarge Knowledge Base, \genk}\\
\multirow{4}{*}{WordNet} & \thead{hyper$(x, y)$} & & \esym \\
& anto(x, y) & & \csym \\
& syno(x, y) &  \makecell[c]{Replace $x$ \\with $y$ in $s$}  & \esym \\
\cline{1-2} \cline{4-4}
\T PPDB & $x \equiv y$ &  & \esym \\
\cline{1-2} \cline{4-4}
\T SICK & $c(x, y)$ &  & $c$ \\
% \hline
% \multicolumn{4}{l}{}\\
% \hline
\rowcolor[gray]{.8}\multicolumn{4}{c}{\Tlarge Hand-authored, \genh}\\
\T Domain knowledge & \textsc{neg} & \textsc{negate}($s$) & \csym \\
% \hline
% \multicolumn{4}{l}{}\\
% \hline
\rowcolor[gray]{.8}\multicolumn{4}{c}{\Tlarge Neural Model, \gens}\\
\T Training data & (s2s, $c$) & $\gens_c(s)$ & $c$ \\
\end{tabular}
\caption{\label{tab:func} Various generators $\gen_\rho$ characterized by their source, (partial) transformation function $f_\rho$ as applied to a sentence $s$, and entailment label $g_\rho$ }
%\vspace{-3mm}
%\ashish{changed the layout a bit, but the older treatment may have been better}
\end{table}

%%%%%%%%%%%%%%%
\subsection{Knowledge-Guided Generators}
\label{sec:genr}
%%%%%%%%%%%%%%%
Large knowledge-bases such as WordNet and PPDB contain lexical equivalences and other relationships highly relevant for entailment models. However, even large datasets such as SNLI generally do not contain most of these relationships in the training data. E.g., that \emph{two dogs} entails \emph{animals} isn't captured in the SNLI data. We define simple generators based on lexical resources to create adversarial examples that capture the underlying knowledge. This allows models trained on these examples to learn these relationships.

As discussed earlier, there are different ways of incorporating such symbolic knowledge into neural models. Unlike task-agnostic ways of approaching this goal from a word embedding perspective~\cite{faruqui2014retrofitting,mrkvsic2016counter} or the model-specific approach~\cite{sha2016recognizing,chen2018natural}, we use this knowledge to generate task-specific examples. This allows any entailment model to learn how to use these relationships \emph{in the context of the entailment task}, helping them outperform the above task-agnostic alternative.

Our knowledge-guided example generators, $\genk_\rho$, use lexical relations available in a knowledge-base: $\rho:=r(x, y)$ where the relation $r$ (such as synonym, hypernym, etc.) may differ across knowledge bases. We use a simple (partial) transformation function, $f_\rho(s)$:=``Replace $x$ in $s$ with $y$'', as described in an earlier example. In some cases, when part-of-speech (POS) tags are available, the partial function requires the tags for $x$ in $s$ and in $r(x, y)$ to match. The entailment label $g_\rho$ for the resulting examples is also defined based on the relation $r$, as summarized in Table~\ref{tab:func}.

This idea is similar to Natural Logic Inference or NLI~\cite{lakoff1970linguistics,sommers1982logic,angeli2014naturalli} where words in a sentence can be replaced by their hypernym/hyponym to produce entailing/neutral sentences, depending on their context. We propose a context-agnostic use of lexical resources that, despite its simplicity, already results in significant gains. We use three sources for generators:

\paragraph{WordNet}~\citep{miller1995wordnet}
%DONE(TK) \ashish{Something like: it's a large hand-curated semantic lexicon, from which we use hypernym, synonym, and antonym relations. Similar to PPDB, we do POS-tag based matching and restrict substitutions to nouns and verbs.}
is a large, hand-curated, semantic lexicon with synonymous words grouped into \emph{synsets}. Synsets are connected by many semantic relations, from which we use \textit{hyponym} and \textit{synonym} relations to generate entailing sentences, and \textit{antonym} relations to generate contradicting sentences\footnote{A similar approach was used in a parallel work to generate an adversarial dataset from SNLI~\cite{glockner_acl18}.}. Given a relation $r(x,y)$, the (partial) transformation function $f_\rho$ is the POS-tag matched replacement of $x$ in $s$ with $y$, and requires the POS tag to be noun or verb. NLI provides a more robust way of using these relations based on context, which we leave for future work.

\paragraph{PPDB}~\citep{ganitkevitch2013ppdb}
%DONE(TK) \ashish{simplify to something like: it's a large resource of lexical paraphrases. We use XX paraphrases in PPDB-S as equivalence relations and use POS tag matching between PPDB string and the text to replace for ... (some nice wording for this `detail')}
is a large resource of lexical, phrasal, and syntactic paraphrases. We use 24,273 lexical paraphrases in their smallest set, PPDB-S~\cite{pavlick2015ppdb2}, as equivalence relations, $x \equiv y$. The (partial) transformation function $f_\rho$ for this generator is POS-tagged matched replacement of $x$ in $s$ with $y$, and the label $g_\rho$ is \entails.

\paragraph{SICK}~\citep{marelli2014sick}
%DONE(TK) \ashish{could be shrunk to something like: was created to evaluate the ability of entailment models to capture compositional knowledge, beltagy et al extracted 12,508 e/c/n relations from it (we ignore the positional information for simplicity); we use these for knowledge-guided example generation as discussed above.}
%or Sentences Involving Compositional Knowledge
is dataset with entailment examples of the form $(p, h, c)$, created to evaluate an entailment model's ability to capture compositional knowledge via hand-authored rules. We use the 12,508 patterns of the form
%$F$=$\{$c(x, y)$\}$
$c(x, y)$
extracted by~\citet{beltagy2016representing} by comparing sentences in this dataset, with the property that for each SICK example $(p, h, c)$, replacing (when applicable) $x$ with $y$ in $p$ produces $h$. For simplicity, we ignore positional information in these patterns. The (partial) transformation function $f_\rho$ is replacement of $x$ in $s$ with $y$, and the label $g_\rho$ is $c$.
%They defined rules to automatically create such entailment examples. We used the patterns extracted from the SICK dataset~\cite{beltagy2015representing} as the knowledge base. We use the 12,508 patterns of the form $F$=$\{$(lhs, rhs, c)$\}$\footnote{For simplicity, we ignored the positional information in the patterns.} from their train and test set with one rule template per entailment label, c: $R_{\textit{SICK}, e}(s, f)$: \textit{Replace the lhs by rhs in s.}
%\tushark{Might be overloading the notation too much.}
%We use the 12,508 SICK rules from their train and test set, we use entailment, neutral, and contradiction labels.
%The contradiction rules are then applied by swapping the true labels in our training.
%Remove position tags (e.g., car-7 moves-8)

\subsection{Hand-Defined Generators}
\label{sec:hand}

Even very large entailment datasets have no or very few examples of certain otherwise common linguistic constructs such as negation,\footnote{Only 211 examples (2.11\%) in the SNLI training set contain negation triggers such as not, 'nt, etc.} causing models trained on them to struggle with these constructs. A simple model-agnostic way to alleviate this issue is via a negation example generator whose transformation function $f_\rho(s)$ is \textsc{negate}$(s)$, described below, and the label $g_\rho$ is \contradicts.

\textsc{negate}(s): If $s$ contains a `be' verb (e.g., is, was), add a ``not'' after the verb. If not, also add a ``did'' or ``do'' in front based on its tense. E.g., change ``A person is crossing'' to ``A person is not crossing'' and ``A person crossed'' to ``A person did not cross.'' While many other rules could be added, we found that this single rule covered a majority of the cases.
Verb tenses are also considered\footnote{{https://www.nodebox.net/code/index.php/Linguistics}} and changed accordingly.
Other functions such as dropping adverbial clauses or changing tenses could be defined in a similar manner. 

Both the knowledge-guided and hand-defined generators make local changes to the sentences based on simple rules. It should be possible to extend the hand-defined rules to cover the long tail (as long as they are procedurally definable). However, a more scalable approach would be to extend our generators to trainable models that can cover a wider range of phenomena than hand-defined rules. Moreover, the applicability of these rules generally depends on the context which can also be incorporated in such trainable generators.
%\tushark{Do we change the output tense too?}\dk{I use a library for checking the tense. Then, I convert it to present tense and add ``did'' instead of ``do'',}

\subsection{Neural Generators}
\label{sec:gens}

For each entailment class $c$, we use a trainable sequence-to-sequence neural model~\cite{sutskever2014sequence,luong2015effective} to generate an entailment example $(s, s', c)$ from an input sentence $s$. The seq2seq model, trained on examples labeled $c$, itself acts as the transformation function $f_\rho$ of the corresponding generator $\gens_c$. The label $g_\rho$ is set to $c$. The joint probability of seq2seq model is:
\begin{align}
%\gen_{\orig,c} =  &\mathbb{G} (H | P, c ; \phi)  = \mathbb{G}_c (\mathcal{X}  ; \phi) \\
%\tilde{H} = & \argmax \mathbb{G}_c (\mathcal{X}  ; \phi) \\
%\hat{\phi_{\mathcal{X},c}} = &\argmin_{\phi} L_{\mathbb{G}_{\mathcal{X},c}} (\tilde{H}, H; \phi)
\gens_{c}(\orig_c; \phi_c) & = \gens_{c}(H_c,P_c; \phi_c) \\
&= \Pi_i P(h_{i,c} | p_{i,c}; \phi_c) P(h_i)
\end{align}

% \tushark{Needs a better explanation of the equations below}
% \ashish{notation seems confusing}
%\mathbb{D}_{\mathcal{X}} = &\mathbb{D} (c | P, H; \theta)  =  \mathbb{D} (\mathcal{X}; \theta)\\
%\tilde{C}  =& \argmax_C \mathbb{D}_{\theta}(\orig) \\
%\hat{\theta}    = &\argmin_{\theta} L_{\mathbb{D}_{\mathcal{X}}} (c, \tilde{c}; \theta) 
The loss function for training the seq2seq is:
\begin{align}
\hat{\phi}_c & = \argmin_{\phi_c} L(H_c,\gens_{c}(\orig_c; \phi_c)) 
%\synth_{\gens} & = \cup_{c \in \classes} \gens_c(\orig; \hat{\phi}_c)
\end{align}
where
%$L_{\mathbb{G}_{\mathcal{X},c}}$
$L$ is the cross-entropy loss between the original hypothesis $H_c$ and the predicted hypothesis. 
Cross-entropy is computed for each predicted word $w_i$ against the same in $H_c$ given the sequence of previous words in $H_c$
%\tushark{I am assuming the previous words are from $H_c$ and not the predictions during training}
. $\hat{\phi_c}$ are the optimal parameters in $\gens_c$ that minimize the loss for class $c$. We use the single most likely output to generate sentences in order to reduce decoding time.

\subsection{Example Generation}
The generators described above are used to create new entailment examples from the training data. For each example $(p, h, c)$ in the data, we can create two new examples: $\left(p, f_\rho(p), g_\rho\right)$ and $\left(h, f_\rho(h), g_\rho\right)$.

\begin{figure}[t]
  \centering
  {
  \includegraphics[trim=8.6cm 5.79cm 6.1cm 3.4cm,clip,width=.98\linewidth]{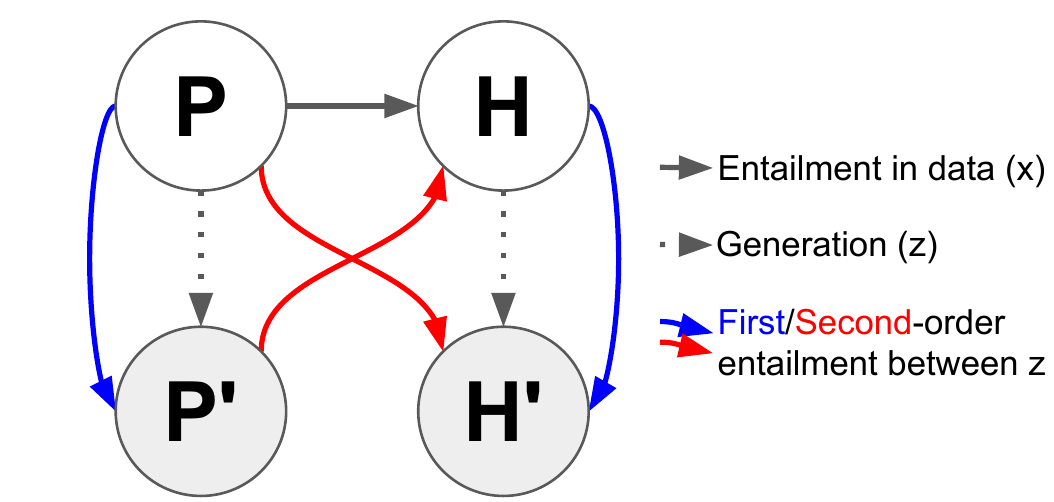}
  }
  \caption{\label{fig:reasoning} Generating first-order (blue) and second-order (red) examples.}
  %\vspace{-2mm}
\end{figure}

The examples generated this way using \genk and \genh can, however, be relatively easy, as the premise and hypothesis would differ by only a word or so. We therefore compose such simple (``first-order'') generated examples with the original input example to create more challenging ``second-order'' examples. 
We can create second-order examples by composing the original example $(p, h, c)$ with a generated sentence from hypothesis, $f_\rho(h)$ and premise, $f_\rho(p)$.
Figure~\ref{fig:reasoning} depicts how these two kinds of examples are generated from an input example $(p, h, c)$.

First, we consider the second-order example between the original premise and the transformed hypothesis: $(p, f_\rho(h), \bigoplus(c, g_\rho))$, where $\bigoplus$, defined in the left half of Table~\ref{tab:second}, composes the input example label $c$ (connecting $p$ and $h$) and the generated example label $g_\rho$ to produce a new label. For instance, if $p$ entails $h$ and $h$ entails $f_\rho(h)$, $p$ would entail $f_\rho$. In other words, $\bigoplus(\esym, \esym)$ is $\esym$. 
%The composed label is frequently \neutral, and in two cases undetermined (denoted `?'). 
 For example, composing (``A man is playing soccer'', ``A man is playing a game'', \esym) with a generated hypothesis $f_{\rho}(h)$: ``A person is playing a game.'' will give a new second-order entailment example: (``A man is playing soccer'', ``A person is playing a game'', \esym).

%We use the same join relation to describe the $\bigoplus$ function with few changes. Since ~\namecite{icard2014recent} have additional relations that are not captured by our entailment classes, we ignore compositions that lead to such relations (shown with ? in Table~\ref{tab:second})\footnote{We map the exclusivity/alternation relation to contradiction.}.

\begin{table}[t]
\fontsize{10.1}{11}\selectfont
    \centering
    \setlength{\tabcolsep}{0.7ex}
    \begin{tabular}{@{}cc|c@{\hskip 3ex}cc|c@{}}
    \hline
$\bm{p} \Rightarrow h$ & $h \Rightarrow \bm{h'}$ & $\bm{p} \Rightarrow \bm{h'}$ & $p \Rightarrow \bm{h}$ & $p \Rightarrow \bm{p'}$ & $\bm{p'} \Rightarrow \bm{h}$ \\\hline
%l1 & l2 & l3 & l1 & l2 & l3 \\ \hline\hline
$c$ & $g_\rho$ & $\bigoplus$ & $c$ & $g_\rho$ & $\bigotimes$ \\%\hline
\rowcolor{gray!40}\T \esym & \esym & \esym & \esym & \esym & ?\\
\rowcolor{gray!40}\R \esym & \csym & \csym & \esym & \csym & ? \\
\rowcolor{gray!40}\R \esym & \nsym & \nsym & \esym & \nsym & \nsym \\
\rowcolor{gray!30}\R \csym & \esym & ? & \csym & \esym & ? \\
\rowcolor{gray!30}\R \csym & \csym & ? & \csym & \csym & ? \\
\rowcolor{gray!30}\R \csym & \nsym & \nsym & \csym & \nsym & \nsym \\
\rowcolor{gray!10}\R \nsym & \esym & \nsym & \nsym & \esym & \nsym \\
\rowcolor{gray!10}\R \nsym & \csym & \nsym & \nsym & \csym & \nsym \\
\rowcolor{gray!10}\B \nsym & \nsym & \nsym & \nsym & \nsym & \nsym  \\\hline
      \end{tabular}
      \caption{\label{tab:second}Entailment label composition functions $\bigoplus$ (left) and $\bigotimes$ (right) for creating second-order examples. $c$ and $g_\rho$ are the original and generated labels, resp. \esym: \entails, \csym: \contradicts, \nsym: \neutral, ?: undefined}
      %\vspace{-1mm}
      %$?$ is used to indicate cases where the composition is not defined by the three entailment classes.}
\end{table}

Second, we create an example from the generated premise to the original hypothesis: $(f_\rho(p), h, \bigotimes(g_\rho, c))$. The composition function here, denoted $\bigotimes$ and defined in the right half of Table~\ref{tab:second}, is often undetermined. For example, if $p$ entails $f_\rho(p)$ and $p$ entails $h$, the relation between $f_\rho(p)$ and $h$ is undetermined i.e. $\bigotimes(\esym, \esym)=?$. While this particular composition $\bigotimes$ often leads to undetermined or neutral relations, we use it here for completeness. For example, composing the previous example with a generated 
\neutral premise, $f_\rho(p)$: ``A person is wearing a cap'' would generate an example (``A person is wearing a cap'', ``A man is playing a game'', \nsym)

The composition function $\bigoplus$ is the same as the ``join'' operation in natural logic reasoning~\cite{icard2014recent}, except for two differences: (a) relations that do not belong to our three entailment classes are mapped to `?', and (b) the exclusivity/alternation relation is mapped to \contradicts. The composition function $\bigotimes$, on the other hand, does not map to the join operation.

\begin{figure*}[t]
\centering
\includegraphics[trim=3.4cm 5.0cm 2.9cm 2.8cm,clip,width=.99\linewidth]{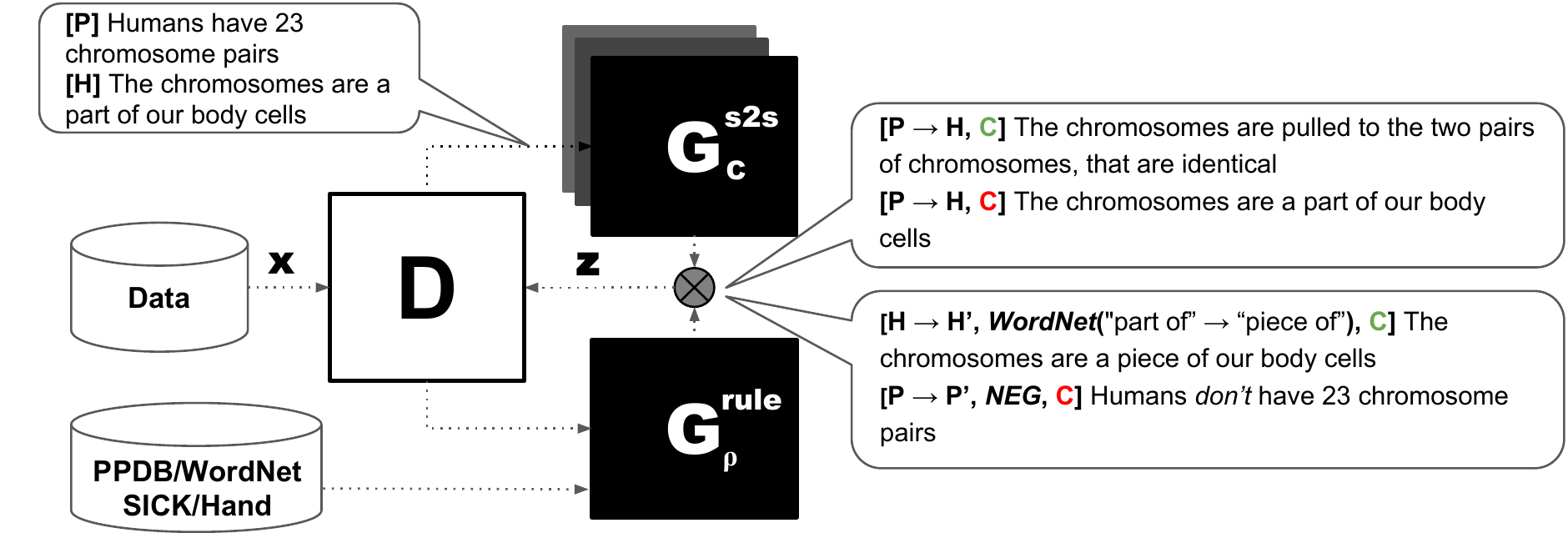}
\caption{\label{fig:architecture} Overview of \method, our model for knowledge-guided textual entailment.}
%\tushark{Needs an update; no more swap. Also different model names.}\dk{done.}
\end{figure*}

\subsection{Implementation Details}
\label{sec:implementation-details}
Given the original training examples \orig, we generate the examples from each premise and hypothesis in a batch using \genk and \genh. We also generate new hypothesis per class for each premise using $\gens_c$. Using all the generated examples to train the model would, however, overwhelm the original training set. For examples, our knowledge-guided generators \genk can be applied in 17,258,314 different ways. 

To avoid this, we sub-sample our synthetic examples to ensure that they are proportional to the input examples $X$, specifically they are bounded to $\alpha|X|$ where $\alpha$ is tuned for each dataset. Also, as seen in Table~\ref{tab:second}, our knowledge-guided generators are more likely to generate \neutral examples than any other class. To make sure that the labels are not skewed, we also sub-sample the examples to ensure that our generated examples have the same class distribution as the input batch. The SciTail dataset only contains two classes: \entails mapped to \esym and \neutral mapped to \csym. As a result, generated examples that do not belong to these two classes are ignored. 

The sub-sampling, however, has a negative side-effect where our generated examples end up using a small number of lexical relations from the large knowledge bases. On moderate datasets, this would cause the entailment model to potentially just memorize these few lexical relations.  Hence, we generate new entailment examples for each mini-batch and update the model parameters based on the training+generated examples in this batch.

The overall example generation procedure goes as follows: For each mini-batch $X$ (1) randomly choose 3 applicable rules per source and sentence  (e.g., replacing “men” with “people” based on PPDB in premise is one rule), (2) produce examples $Z_{all}$ using $\genk$, $\genh$ and \gens, (3) randomly sub-select examples $Z$ from $Z_{all}$ to ensure the balance between classes and $|Z| = \alpha |X|$.

%%%%%%%%%%%%%%%%%%%%%%%%%%%%%%%%%%%%%%%%%%%%
\section{\method}
%%%%%%%%%%%%%%%%%%%%%%%%%%%%%%%%%%%%%%%%%%%%

%\tushark{Update figure and description.} 
Figure~\ref{fig:architecture} shows the complete architecture of our model, \method(ADVersarial training for textual ENTailment Using Rule-based Examples.). The entailment model \disc\ is shown with the white box and two proposed generators are shown using black boxes. We combine the two symbolic untrained generators, \genk and \genh into a single \genr model. We combine the generated adversarial examples \synth\ with the original training examples \orig\ to train the discriminator. Next, we describe how the individual models are trained and finally present our new approach to train the generator based on the discriminator's performance.

\subsection{Discriminator Training}
We use one of the state-of-the-art entailment models (at the time of its publication) on SNLI, decomposable attention model~\cite{parikh2016decomposable} with intra-sentence attention as our discriminator \disc.
The model attends each word in hypothesis with each word in the premise, compares each pair of the attentions, and then aggregates them as a final representation. This discriminator model can be easily replaced with any other entailment model without any other change to the \method\ architecture.
We pre-train our discriminator \disc\ on the original dataset, X=(P, H, C) using:
\begin{align}
\disc(\orig; \theta) & = \argmax_{\hat{C}} \disc(\hat{C} | P, H; \theta) \\
%\mathbb{D}_{\mathcal{X}} = &\mathbb{D} (c | P, H; \theta)  =  \mathbb{D} (\mathcal{X}; \theta)\\
%\tilde{C}  =& \argmax_C \mathbb{D}_{\theta}(\orig) \\
%\hat{\theta}    = &\argmin_{\theta} L_{\mathbb{D}_{\mathcal{X}}} (c, \tilde{c}; \theta) 
\hat{\theta}  = &\argmin_{\theta} L(C,\disc(\orig; \theta)) 
\end{align}
%where $H$ and $P$ are hypothesis and premise sentences in the original dataset \orig, 
%$\tilde{C}$ is the set of predicted classes for this dataset,  
 where $L$ is cross-entropy loss function between the true labels, $Y$ and the predicted classes, and $\hat{\theta}$ are the learned parameters. %optimal parameters for $\mathbb{D}$ that minimize the discriminator loss.
%\dk{define $\mathbb{D}_{\mathcal{Z}}$ for adversarial part later section}

\subsection{Generator Training}
Our knowledge-guided and hand-defined generators are symbolic parameter-less methods which are not currently trained. For simplicity, we will refer to the set of symbolic rule-based generators as $\genr := \genk \cup \genh$. The neural generator \gens, on the other hand, can be trained as described earlier. We leave the training of the symbolic models for future work.

%%%%%%%%%%%%%%%
\subsection{Adversarial Training}
%%%%%%%%%%%%%%%
We now present our approach to iteratively train the discriminator and generator in a GAN-style framework. Unlike traditional GAN~\cite{goodfellow2014generative} on image/text generation that aims to obtain better generators, our goal is to build a robust discriminator regularized by the generators (\gens and \genr).
The discriminator and generator are iteratively trained against each other to achieve better discrimination on the augmented data from the generator and better example generation against the learned discriminator. Algorithm~\ref{alg:inference} shows our training procedure.

\begin{algorithm}[t]
%\footnotesize
\caption{\label{alg:inference} Training procedure for \method. }
%The class $c$ has \textit{e}ntailment, \textit{c}ontradiction, and \textit{n}eutral. The rules $r$ has \textit{p}pdb, \textit{s}ick, \textit{w}ordnet, and \textit{h}and. $\textbf{X}$ and $\mathcal{Z}$ are original and generated data, respectively.}

\begin{algorithmic}[1]

\State \texttt{pretrain} discriminator $\mathbb{D}({\hat{\theta}})$ on $\textbf{X}$; 
\State \texttt{pretrain} generators $\gens_c (\hat{\phi})$ on $\textbf{X}$;
\For{ number of training iterations }
	\For{\texttt{mini-batch $B \leftarrow \orig$ }}
	   \State \texttt{generate} examples from \gen
       \Indent
       \State $\synth_{G}$$\Leftarrow$$\gen(B ; \phi)$,
       \EndIndent
       %\State \texttt{generate} examples from \genr
       %\Indent
       %\State $\synth_{R}$$\Leftarrow$$\genr(B ; \gamma)$,
       %\EndIndent
      %\State \texttt{generate} 1st-order sentences:
      %\Indent 
      %\State $\mathcal{Z}_{G}^{1}$$\Leftarrow$$\mathbb{G}(\mathcal{X}_p ; \phi)$, $\mathcal{Z}_{R}^{1}$ $\Leftarrow$$\mathbb{R}(\mathcal{X};\gamma)$;
      %\EndIndent
      %\State \texttt{generate} 2nd-order sentences:
      %\Indent 
      %\State    $\mathcal{Z}_{G}^{2}$$\Leftarrow$$\mathbb{G}(\mathcal{Z}_{R}^{1};\phi)$, $\mathcal{Z}_{R}^{2}$ $\Leftarrow$ $\mathbb{R}(\mathcal{Z}_{G}^{1};\gamma)$;
      %\EndIndent

	  \State \texttt{balance} $\orig$ and $\synth_{G}$ s.t. $|\synth_G|$ $\leq \alpha |\orig|$
      \State \texttt{optimize} discriminator:
      \Indent 
        \State $\hat{\theta} = \argmin_{\theta} L_{\mathbb{D}} (\orig+\synth_{G}; \theta)$ 
      \EndIndent
      \State \texttt{optimize} generator:
      \Indent 
        \State $\hat{\phi} = \argmin_{\phi} L_{\gens} (\mathcal{Z}_{G}; L_\disc;\phi)$
        %\State $\hat{\gamma} = \argmin_{\phi} L_{\mathbb{R}} (\mathcal{Z}_{R}; \gamma)$        
      \EndIndent
%       \State \texttt{optimize} $\hat{\phi_{\mathcal{X},c}} = \argmin_{\phi} L_{\mathbb{G}_{\mathcal{X},c}} (\tilde{H}, H; \phi)$
%       \State \texttt{optimize} $\hat{\gamma} = \argmin_{\gamma} L_{\mathbb{R}} (\gamma) = L_{\mathbb{D}} (\mathcal{Z}_{\mathbb{R}} ; \gamma)$

    	\State Update $\theta \leftarrow \hat{\theta} ; \phi \leftarrow \hat{\phi} $
	\EndFor
\EndFor
\end{algorithmic}
\end{algorithm}

%     \State $L_{\mathbb{D}\mathcal{R}\mathcal{Z}} = \underset{\mathbb{D}}{\text{max}} E_{\mathcal{X} \sim P_{r,g}} \mathbb{D}(x) - E_{\mathcal{X} \sim p(\mathcal{Z})} \mathbb{D}(\mathbb{G}(\mathcal{Z})) - E_{\mathcal{X} \sim p(\mathcal{R})} \mathbb{D}(\mathbb{R}(\mathcal{Z})) $ 

%Now, we combine all individual modules -- generators, \rulenet, and discriminator -- in one place, and learn them together as an end-to-end training.

First, we pre-train the discriminator $\mathbb{D}$ and the seq2seq generators $\gens$ on the original data $\orig$.
%, and \rulenet $\mathbb{R}$\tushark{is this pre-trained too?}\dk{It's optional but let's see whether train version is better than non-train version.} on the original data $\orig$.
%We set then a number of iteration to train our algorithms~\footnote{number of iteration is our hyper-parameter. We use 30 in our experiment.}.
We alternate the training of the discriminator and generators over K iterations (set to 30 in our experiments).

% \begin{figure}[h]
%   \centering
%   {
%   \includegraphics[trim=4.2cm 1.3cm 4.5cm 1cm,clip,width=.50\linewidth]{figs/{genreasoning2}.pdf}  
%   \vspace{-0mm}
%   }
%   \caption{\label{fig:reasoning2} Generating first or second order sentences from \rulenet $\mathcal{R}$ and  generators $\mathcal{G}$.}
% \end{figure}

For each iteration, we take a mini-batch $B$ from our original data $\orig$. For each mini-batch, we generate new entailment examples, $\synth_G$ using our adversarial examples generator.
%first-order and second-order entailment examples, $\synth_G$ and $\synth_R$ from the seq2seq \gens and rule-based \genr respectively.
%We can generate entailment, neutral, or contradict sentences $\mathcal{Z}_{G}$ given premise using generator first.
%On the other side, we can generate entailment, neutral, or contradict sentences $\mathcal{Z}_{R}$ with rule reasoning described in the earlier section.
%Those sentences are first-orderly generated sentences: $\mathcal{Z}_{G}^1$ and $\mathcal{Z}_{R}^1$.
%Further, we can apply rules to the generator on $\mathcal{Z}_{R}^1$ or apply generators on $\mathcal{Z}_{G}^1$ called second-orderly generated sentences: $\mathcal{Z}_{G}^2$ and $\mathcal{Z}_{R}^2$.
%Look at Figure~\ref{fig:reasoning2} for details.
 Once we collect all the generated examples, %$\mathcal{Z}_{G}^1$, $\mathcal{Z}_{R}^1$, $\mathcal{Z}_{G}^2$ and $\mathcal{Z}_{R}^2$, and the original sentences $\mathcal{X}$, 
we balance the examples based on their source and label (as described in Section~\ref{sec:implementation-details}).
In each training iteration, we optimize the discriminator against the augmented training data, $\orig + \synth_{G}$ and use the discriminator loss to guide the generator to pick challenging examples. For every mini-batch of examples $\orig + \synth_{G}$, we compute the discriminator loss $L(C; \disc(\orig + \synth_{G}; \theta))$ and apply the negative of this loss to each word of the generated sentence in \gens. In other words, the discriminator loss value replaces the cross-entropy loss used to train the seq2seq model
%To train the generator, we use the loss of the discriminator on the augmented dataset, $L_{\disc}$ to guide the generator to pick challenging examples by passing this loss to the decoder in the generator 
(similar to a REINFORCE~\cite{reinforce} reward). \added{This basic approach uses the loss over the entire batch to update the generator, ignoring whether specific examples were hard or easy for the discriminator. Instead, one could update the generator per example based on the discriminator's loss on that example. We leave this for future work.} %This will allow the generator to focus on the specific weaknesses of the discriminator.}
%\tushark{Still vague. In the pre-training, we use the cross-entropy loss between the predicted words and original hypothesis. What is the loss here ? }\dk{Loss function for \gens is same as the original cross entropy between generated hypothesis and original hypothesis. The thing is, we calculate the loss from \disc and update that loss to \gens not calculating the original loss again. It's simple penalty/reward mechanism of current \gens is doing good job or not based on \disc's loss. }\tushark{Is the loss then $L_\gen(X) + L_\disc(Z + X)$? If so, isn't the second term a constant w.r.t. \gen i.e. would result in zero gradients during backprop ? Or do you explicitly update each gradient with $L_\disc$}\dk{See Slack}

%%%%%%%%%%%%%%%%%%%%%%%%%%%%%%%%%%%%%%%%%%%%%%%%%%%%%%%%%%%%%%%%%%%%%%%%%%%%%%%%%%%%%%%%%%%%%%%%
\section{Experiments}
%%%%%%%%%%%%%%%%%%%%%%%%%%%%%%%%%%%%%%%%%%%%%%%%%%%%%%%%%%%%%%%%%%%%%%%%%%%%%%%%%%%%%%%%%%%%%%%%

Our empirical assessment focuses on two key questions: (a) Can a handful of rule templates improve a state-of-the-art entailment system, especially with moderate amounts of training data?  (b) Can iterative GAN-style training lead to an improved discriminator?

To this end, we assess various models on the two entailment \textbf{datasets} mentioned earlier: SNLI (570K examples) and SciTail (27K examples).\footnote{SNLI has a 96.4\%/1.7\%/1.7\% split and SciTail has a 87.3\%/4.8\%/7.8\% split on train, valid, and test sets, resp.} To test our hypothesis that adversarial example based training prevents overfitting in small to moderate training data regimes, we compare model accuracies on the test sets when using 1\%, 10\%, 50\%, and 100\% subsamples of the train and dev sets.
%We hypothesize our adversarial example-based training leads to robust models by preventing overfitting, a key problem in small to moderate training datasets. Hence, to evaluate this robustness, we compare the accuracies of the models on different training set sizes. Specifically, we train models on 1\%, 10\%, 50\%, 100\% of the training and validation sets and evaluate them on the entire test set.

We consider two baseline \textbf{models}: \disc, the Decomposable Attention model~\cite{parikh2016decomposable} with intra-sentence attention using pre-trained word embeddings~\cite{pennington2014glove}; and \discretro which extends \disc with  word embeddings initialized by retrofitted vectors~\cite{faruqui2014retrofitting}. The vectors are retrofitted on PPDB, WordNet, FrameNet, and all of these, with the best results for each dataset reported here.

Our proposed model, \method, is evaluated in three flavors: \disc augmented with examples generated by \genr, \gens, or both, where $\genr = \genk \cup \genh$. In the first two cases, we create new examples for each batch in every epoch using a fixed generator (cf.~Section~\ref{sec:implementation-details}). In the third case (\disc + \genr + \gens), we use the GAN-style training. 

We uses grid search to find the best hyper-parameters for \disc based on the validation set: hidden size 200 for LSTM layer, embedding size 300, 
%vocabulary size 37,287 for SNLI and 24,129 for SciTail \tushark{is this tuned?}\dk{used same hyper-param but almost similar with the best params from NSNet.}\tushark{Overleaf seems to be eating my comment here everytime. My point was that the vocab size is set based on the dataset not tuned right ?}\dk{Slack},
dropout ratio 0.2, and fine-tuned embeddings.
% Figure~\ref{fig:D} shows training (dotted) accuracies on sub-sampled training dataset on $ratio * \orig_{train}$ and testing (solid) accuracies on original test dataset \orig$_{test}$ by \disc.
% The X-axis is the $ratio$ from 0.01, 0.02, 0.03, 0.04, 0.05, 0.1, 0.25, 0.5, 0.75, and 1.0. 

% For SciTail that contains less number of examples than SNLI, it has little fluctuation at first and middle and reaches to the 74.29.
% SNLI shows very stable increases and almost converges at the full ratio=1.0.
% The difference of stableness in test accuracies are mainly because of data size and originality of tasks: SciTail is more likely question answering task which requires external knowledge bases, while SNLI needs surface level of lexical entailment patterns that could be mostly captured by the dataset.

The ratio between the number of generated vs.\ original examples, $\alpha$ is empirically chosen to be 1.0 for SNLI and 0.5 for SciTail, based on validation set performance. Generally, very few generated examples (small $\alpha$) has little impact, while too many of them overwhelm the original dataset resulting in worse scores (cf.~Appendix for more details).

%%%%%%%%%%%%%%%
%\subsection{Adversarial Training}
\subsection{Main Results}
%%%%%%%%%%%%%%%

Table~\ref{tab:augment} summarizes the test set accuracies of the different models using various subsampling ratios for SNLI and SciTail training data.

\begin{table}[tb]
\fontsize{10}{11}\selectfont
\setlength{\doublerulesep}{\arrayrulewidth}
\caption{\label{tab:augment} Test accuracies with different subsampling ratios on SNLI (top) and SciTail (bottom). }
\centering
%\vspace{-3mm}
\begin{tabular}{@{}l|cccc@{}}
\textbf{SNLI}  & 1\%   & 10\% &  50\% & 100\%\\ \hline \hline
% \midrule
\T \disc		& 57.68 & 75.03 &  82.77 &  84.52 \\
\discretro		& 57.04 & 73.45 &  81.18 & 84.14\\
\method\\
$\;\llcorner$ \disc + \gens 		& 58.35 &75.66 &  82.91 & \textbf{84.68}\\
$\;\llcorner$ \disc + \genr 		& \textbf{60.45} & \textbf{77.11} & \textbf{83.51} & 84.40 \\
$\;\llcorner$ \disc + \genr + \gens & 59.33& 76.03 & 83.02 & 83.25\\
\bottomrule
\end{tabular}

\vspace{2ex}
\fontsize{10}{11}\selectfont
\begin{tabular}{@{}l|cccc@{}}
\textbf{SciTail}  & 1\%   & 10\% &  50\% & 100\%\\ \hline \hline
% \midrule
\T \disc		&  56.60 & 60.84 &  73.24 &  74.29\\
\discretro		& 59.75 & 67.99 & 69.05 &  72.63 \\
\method\\
$\;\llcorner$ \disc + \gens 		&\textbf{65.78} & \textbf{70.77} & 74.68 & 76.92   \\
$\;\llcorner$ \disc + \genr 		& 61.74 &  66.53 & 73.99 & \textbf{79.03} \\
$\;\llcorner$ \disc + \genr + \gens & 63.28 & 66.78& \textbf{74.77} & 78.60\\
\hline
\end{tabular}
%\vspace{-3mm}
\end{table}

We make a few observations. First, \discretro is ineffective or even detrimental in most cases, except on SciTail when 1\% (235 examples) or 10\% (2.3K examples) of the training data is used. The gain in these two cases is likely because retrofitted lexical rules are helpful with extremely less data training while not as data size increases.

% \ashish{add some hypothesis here. then describe some salient positive observations involving \method.}
On the other hand, our method always achieves the best result compared to the baselines (\disc and \discretro).
Especially, significant improvements are made in less data setting: +2.77\% in SNLI (1\%) and 9.18\% in SciTail (1\%). Moreover, \disc + \genr's accuracy on SciTail (100\%) also outperforms the previous state-of-the-art model (DGEM~\cite{khot2018scitail}, which achieves 77.3\%) for that dataset by 1.7\%.

Among the three different generators combined with \disc, both \genr and \gens are useful in SciTail, while \genr is much more useful than \gens on SNLI.
We hypothesize that seq2seq model trained on large training sets such as SNLI  will be able to reproduce the input sentences. Adversarial examples from such a model are not useful since the entailment model uses the same training examples. However, on smaller sets, the seq2seq model would introduce noise that can improve the robustness of the model. % while it helps generalize the model with less data such SciTail.\tushark{Counters our previous pt that s2s are unreliable on smaller datasets.}

\begin{table}[tb]
\caption{\label{tab:ablation_rulegen} Test accuracies across various rules \rules and classes \classes. Since SciTail has two classes, we only report results on two classes of \gens}
\centering
\fontsize{10.5}{11}\selectfont
\setlength{\doublerulesep}{\arrayrulewidth}
\begin{tabular}{@{}l|l|l|l@{}}
% \toprule
&\rules/\classes & SNLI (5\%) & SciTail (10\%) \\\hline \hline
% \midrule
\parbox[t]{2mm}{\multirow{5}{*}{\rotatebox[origin=c]{90}{\disc+\genr}}}
&\T \disc & 69.18 & 60.84 \\
&+ PPDB		&	\textbf{72.81} (\textbf{+3.6\%})& 65.52 (+4.6\%)\\
&+ SICK		&	71.32 (+2.1\%)& 67.49 (+6.5\%)\\
&+ WordNet 	&	71.54 (+2.3\%)&	64.67 (+3.8\%)\\
&+ HAND		&	71.15 (+1.9\%)& \textbf{69.05} (\textbf{+8.2\%})\\
&+ all		&	71.31 (+2.1\%)& 64.16 (+3.3\%)\\
\hline
\parbox[t]{2mm}{\multirow{5}{*}{\rotatebox[origin=c]{90}{\disc+\gens}}}
&\T \disc 			& 69.18	& 60.84 \\
&+ positive		& 71.21 (+2.0\%)	&67.49 (+6.6\%)\\
&+ negative		& 71.76 (+2.6\%)	&68.95 (+8.1\%)\\
&+ neutral		& 71.72 (+2.5\%)	&-\\
&+ all			& \textbf{72.28} (\textbf{+3.1\%})	&	\textbf{70.77} (\textbf{+9.9\%})\\
\hline
\end{tabular}
\end{table}

\begin{table*}[ht] %{\linewidth}
\fontsize{10.5}{11}\selectfont
%\vspace{-2mm}
\centering
\renewcommand{\arraystretch}{1.5}
\caption{\label{tab:generated} Given a premise \textbf{P} (underlined), examples of hypothesis sentences \textbf{H'} generated by seq2seq generators \gens, and premise sentences \textbf{P'} generated by rule based generators \genr, on the full SNLI data. Replaced words or phrases are shown in \textbf{bold}. \added{This illustrates that even simple, easy-to-define rules can generate useful adversarial examples.}
%\tushark{Statement from our rebuttal: Our goal here was to show that a system with simple, easy-to-define rules can generate useful adversarial examples that lead to more robust models. To further support this claim, we will improve the analysis of the generated examples, especially focusing on the quality and coverage of the examples.}\ashish{added the blue sentence. does it suffice to clarify the purpose of this table? also updated the main caption text.}
}
%\vspace{-2mm}
\begin{tabular}{@{}>{\arraybackslash}l|l@{}}
\Xhline{2\arrayrulewidth}
\hline
\textbf{P} &\underline{a person on a horse jumps over a broken down airplane} \\
\textbf{H'}: $\mathbb{G}^{\mathrm{s2s}}_{c=\esym}$  &  a person is on a horse jumps over a rail, a person jumping over a plane\\
\textbf{H'}: $\mathbb{G}^{\mathrm{s2s}}_{c=\csym}$ &  a person is riding a horse in a field with a dog in a red coat\\
\textbf{H'}: $\mathbb{G}^{\mathrm{s2s}}_{c=\nsym}$ & a person is in a blue dog is in a park\\
\hline
\textbf{P} (or \textbf{H}) & \underline{a dirt bike rider catches some air going off a large hill}\\
\textbf{P'}: $\mathbb{G}^{\mathrm{KB(PPDB)}}_{\rho = \equiv, g_\rho=\esym}$ &  a dirt \textbf{motorcycle} rider catches some air going off a large hill\\
\textbf{P'}: $\mathbb{G}^{\mathrm{KB(SICK)}}_{\rho = c, g_\rho=\nsym}$  & a dirt bike \textbf{man on yellow bike} catches some air going off a large hill\\
\textbf{P'}: $\mathbb{G}^{\mathrm{KB(WordNet)}}_{\rho = syno, g_\rho=\esym}$ & a dirt bike rider catches some \textbf{atmosphere} going off a large hill \\
\textbf{P'}: $\mathbb{G}^{\mathrm{Hand}}_{\rho=\textsc{neg},g_\rho=\csym}$ & a dirt bike rider \textbf{do not catch} some air going off a large hill\\
\hline
\end{tabular}
%\vspace{-2mm}
\end{table*}

%%%%%%%%%%%%%%%
\subsection{Ablation Study}
%%%%%%%%%%%%%%%
To evaluate the impact of each generator, we perform ablation tests against each symbolic generator in \disc + \genr and the generator $\gens_c$ for each entailment class $c$. We use a 5\% sample of SNLI and a 10\% sample of SciTail. The results are summarized in Table~\ref{tab:ablation_rulegen}.

Interestingly, while PPDB (phrasal paraphrases) helps the most (\textbf{+3.6\%}) on SNLI, simple negation rules help significantly (\textbf{+8.2\%}) on SciTail dataset. Since most entailment examples in SNLI are minor rewrites by Turkers, PPDB often contains these simple paraphrases. For SciTail, the sentences are authored independently with limited gains from simple paraphrasing. However, a model trained on only 10\% of the dataset (2.3K examples) would end up learning a model relying on purely word overlap. We believe that the simple negation examples introduce \neutral examples with high lexical overlap, forcing the model to find a more informative signal.

%Our ablation tests (not reported here) also indicate that applying \genr to both premise and hypothesis is better than applying it to only one and that generating both first- and second-order examples is generally better than generating only one kind in most cases.\tushark{still true?}\dk{Overall yes but there are also few runs when specific order was better. Let's remove this one. }
%, or premise only or hypothesis only, or whether to generate first-order examples only or second-order examples only. It turns out applying rules to both premise and hypothesis, and generating both first and second order examples works the best. 

On the other hand, using all classes for \gens results in the best performance, supporting the effectiveness of the GAN framework for penalizing or rewarding generated sentences based on \disc's loss. Preferential selection of rules within the GAN framework remains a promising direction.

%%%%%%%%%%%%%%%
\subsection{Qualitative Results}
%%%%%%%%%%%%%%%

Table~\ref{tab:generated} shows examples generated by various methods in \method. As shown, both seq2seq and rule based generators produce reasonable sentences according to classes and rules. As expected, seq2seq models trained on very few examples generate noisy sentences. The quality of our knowledge-guided generators, on the other hand, does not depend on the training set size and they still produce reliable sentences.

%%%%%%%%%%%%%%%
\subsection{Case Study: Negation}
%%%%%%%%%%%%%%%

For further analysis of the negation-based generator in Table~\ref{tab:example}, 
we collect only the negation examples in test set of SNLI, henceforth referred to as \negasnli.
Specifically, we extract examples where either the premise or the hypothesis contains ``not'', ``no'', ``never'', or a word that ends with ``n't'.
These do not cover more subtle ways of expressing negation such as ``seldom'' and the use of antonyms.
%filters may have high recall but probably less precision because negation also happens in other lexical usage (e.g., ``seldom'', antonyms). 
\negasnli contains 201 examples with the following label distribution: 51 (25.4\%) neutral, 42 (20.9\%) entails, 108 (53.7\%) contradicts.
% It is natural that contradiction is highly correlated to negation patterns. 
Table~\ref{tab:nega_example} shows examples in each category.

\begin{table}[ht]
\fontsize{10.5}{11}\selectfont
\caption{\label{tab:nega_example} Negation examples in \negasnli}%: premise (top) and hypothesis (bottom).}
\centering
\begin{tabularx}{\linewidth}{l|l}
\hline
\esym	& \makecell[l]{P: several women are playing volleyball.\\
H: this doesn't look like soccer.}\\\hline
\nsym	& \makecell[l]{P: a man with no shirt on is performing \\with a baton.\\
H: a man is trying his best at the national \\
championship of baton.}\\\hline
\csym	& \makecell[l]{P: island native fishermen reeling in their \\
nets after a long day's work.	\\
H: the men did not go to work today but \\
instead played bridge.}	\\
\hline
\end{tabularx}
\end{table}

While \disc achieves an accuracy of only 76.64\%\footnote{This is much less than the full test accuracy of 84.52\%.} on \negasnli, \disc + \genh with \textsc{negate} is substantially more successful (+6.1\%) at handling negation, achieving an accuracy of 82.74\%.

%%%%%%%%%%%%%%%%%%%%%%%%%%%%%%%%%%%%%%%%%%%%%%%%
\section{Conclusion}
%%%%%%%%%%%%%%%%%%%%%%%%%%%%%%%%%%%%%%%%%%%%%%%%
We introduced an adversarial training architecture for textual entailment. Our seq2seq and knowledge-guided example generators, trained in an end-to-end fashion, can be used to make any base entailment model more robust. The effectiveness of this approach is demonstrated by the significant improvement it achieves on both SNLI and SciTail, especially in the low to medium data regimes. \added{Our rule-based generators can be expanded to cover more patterns and phenomena, and the seq2seq generator extended to incorporate per-example loss for adversarial training.}

%\subsubsection*{Acknowledgments}

\bibliographystyle{acl_natbib}
\bibliography{adversarial}

% \clearpage
\clearpage

\begin{appendix}
% \section{Appendix}

\section{Rules and Examples}
%\colorbox{red!30}{
\begin{table}[ht]
\centering
\renewcommand{\arraystretch}{1.3}
\caption{\label{tab:rules_examples} Number of rules in \genk}
%Our hand and primitive rules. {\color{red}{$\bm{\neg}$}} means swapping the label}
%\begin{tabular}{cccc}
\begin{tabularx}{\linewidth}{@{}>{\hsize=.08\hsize}c@{}|@{}>{\hsize=.34\hsize}C@{}|@{}>{\hsize=.31\hsize}C@{}|@{}>{\hsize=.27\hsize}C@{}}
%||
%\Xhline{2\arrayrulewidth}
%\hline
%\textbf{\makecell{Hand}} & \multicolumn{3}{c}{{\color{red}{$\bm{\neg}$}}NEGATION, SWAP, DELETE} \\\hline
\textbf{\makecell{KB}}& \textsc{PPDB} & \textsc{SICK} & \textsc{WordNet}\\\hline
\#Rules & 6,977,679 & 12,511 & $\sim$116,000\\\hline
%After filtering & 839,264 & 12,511 & -\\\hline
%LHS & 562,888 & 5,443 & -\\\hline
%Filtered & 1,097,578 & 1,156\\\hline
Examples & because of $\Rightarrow$ due to, wish $\Rightarrow$ would like & woods $\Rightarrow$ wooden area, kid $\nRightarrow$ woman & car $\Rightarrow$ cabin car, hate $\nRightarrow$ love\\\hline
%\Xhline{2\arrayrulewidth}
%\makecell{Types} & Lexical (24,273), O2M (44,590), Phrasal (535,745), NoCCG (234,656)&  Entail (3,955), Neutral (8,378), {\color{red}{$\bm{\neg}$}}Contradict (175) &  Up, Down, PartOf, Synonym, Parallel, {\color{red}{$\bm{\neg}$}}Antonym \\
%\Xhline{2\arrayrulewidth}\Xhline{2\arrayrulewidth}
%\hline
%\Xhline{2\arrayrulewidth}
\end{tabularx}
%\end{tabular}
\end{table}
Table~\ref{tab:rules_examples} shows the number of rules and additional examples for \genk.

\section{Training data sizes}

\begin{figure}[ht]
\centering\vspace{-3mm}
\subfloat[\small{SciTail}]{\includegraphics[trim={10mm 17mm 7mm 20mm},clip,width=\linewidth]{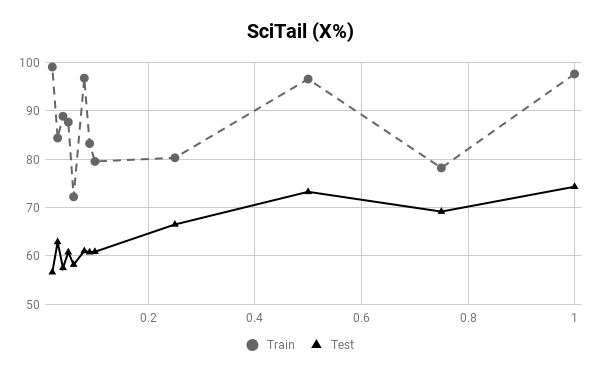}}

\subfloat[\small{SNLI}]{\includegraphics[trim={10mm 17mm 7mm 20mm},clip,width=\linewidth]{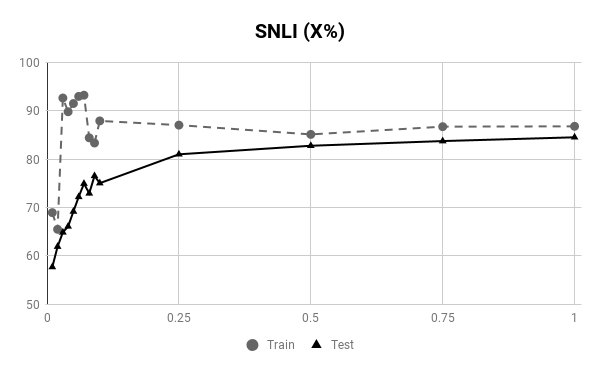}}\vspace{-3mm}
\caption{\disc for SciTail and SNLI. }  \label{fig:D}\vspace{-3mm}
\end{figure}

Figure~\ref{fig:D} shows training (dotted) accuracies on sub-sampled training datasets and testing (solid) accuracies on original test dataset \orig$_{test}$ of \disc over different sub-sampling percentages of the training set.
% The X-axis is the $ratio$ from 0.01, 0.02, 0.03, 0.04, 0.05, 0.1, 0.25, 0.5, 0.75, and 1.0. 
Since SciTail (27K) is much smaller than SNLI (570K), SciTail fluctuates a lot at smaller sub-samples while SNLI  converges with just 50\% of the examples.% has little fluctuation at first and middle and reaches to the 74.29.
% SNLI shows very stable increases and almost converges at the full ratio=1.0.
%The difference of stableness in test accuracies are mainly because of data size and originality of tasks: SciTail is more likely question answering task which requires external knowledge bases, while SNLI needs surface level of lexical entailment patterns that could be mostly captured by the dataset.

\section{Effectiveness of Z/X Ratio, $\alpha$}

\begin{figure}[ht]
\centering
\subfloat[\small{SciTail (D +R$^{rule}$, 10\%) }]{\includegraphics[trim={10mm 17mm 7mm 20mm},clip,width=\linewidth]{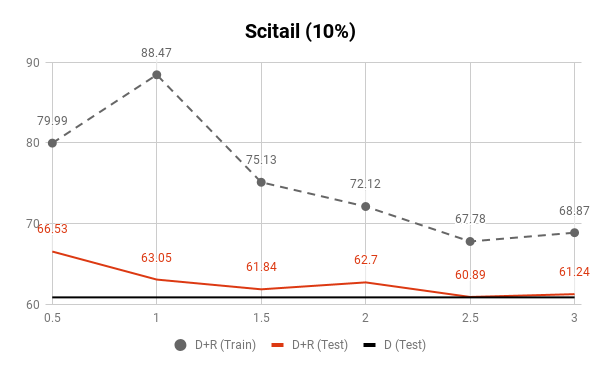}}

\subfloat[\small{SNLI (D +R$^{rule}$, 1\%) }]{\includegraphics[trim={10mm 17mm 7mm 20mm},clip,width=\linewidth]{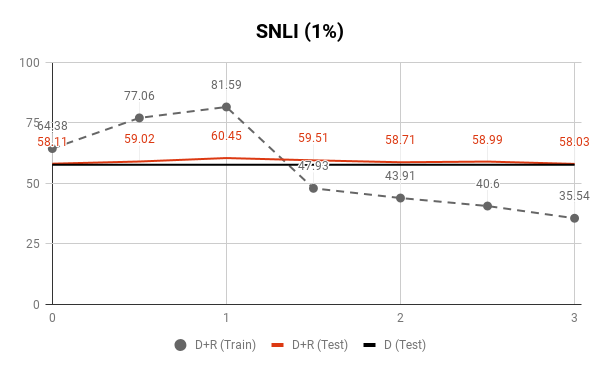}}
\caption{Effect of balancing ratio between $z$ and $x$.  }  \label{fig:ratio}\vspace{-3mm}
\end{figure}
Figure~\ref{fig:ratio} shows train/test accuracies with different balancing ratio between $z$ and $x$.
The dotted line is training accuracies, the solid black horizontal line is testing accuracy of \disc.
The solid red shows test accuracies with different balancing ratio, $\alpha$ (x-axis) from $0.5$, $1.0$, ... $3.0$ from $|z| = \alpha * |x|$ where $|x|$ is fixed as batch size.
The generated examples $z$ are useful up to a point, but the performance quickly degrades for $\alpha > 1.0$ as they overwhelm the original dataset $x$.
% \dk{TODO: Add D+G compared to D+R on one dataset or D+G+R on SNLI/SciTail}

\section{Retrofitting Experiment}
% \begin{figure*}[ht]
% \centering
% \subfloat[\small{D SciTail}]{\includegraphics[trim={0mm 0mm 0mm 0mm},clip,width=0.24\linewidth]{figs/{D_scitail}.png}}
% \subfloat[\small{D SNLI}]{\includegraphics[trim={0mm 0mm 0mm 0mm},clip,width=0.24\linewidth]{figs/{D_snli}.png}}
% \caption{D with different ratio for SciTail and SNLI. }  \label{fig:D}
% \end{figure*}

\begin{table}[ht]
\caption{\label{tab:retrofit} Results of the word vectors retrofitted on different lexicons on each dataset. We pick the best vectors for each task and sub-sampling ratio.}
\centering\vspace{0mm}
\begin{tabular}{r|c|cc}
\toprule
 ratio  & Lexicon & SNLI & SciTail \\\hline
\midrule
1\%	&	framenet	&	56.15 &60.89\\
1\%	&	ppdb	&	\textbf{57.04} &\textbf{62.5}\\
1\%	&	wordnet	&	55.58 &62.2\\
1\%	&	all	&	56.81 &61.14\\
\hline
10\%	&	framenet	&	72.75 &\textbf{67.99}\\
10\%	&	ppdb	&	72.88 &54.74\\
10\%	&	wordnet	&	73.27 &67.29\\
10\%	&	all	&	\textbf{73.45} &66.43\\
\hline
50\%	&	framenet	&	80.95 &66.08\\
50\%	&	ppdb	&	81.14 &67.24\\
50\%	&	wordnet	&	80.62 &\textbf{69.05}\\
50\%	&	all	&	\textbf{81.18} &68.4\\
\hline
100\%	&	framenet	&	83.66 &70.06\\
100\%	&	ppdb	&	\textbf{84.14} &70.16\\
100\%	&	wordnet	&	83.91 &\textbf{72.63}\\
100\%	&	all	&	83.68 &71.12\\
\bottomrule
\end{tabular}\vspace{0mm}
\end{table}
Table~\ref{tab:retrofit} shows the grid search results of retro-fitting vectors~\cite{faruqui2014retrofitting} with different lexical resources. 
To obtain the strongest baseline, we choose the best performing vectors for each sub-sample ratio and each dataset.
Usually, PPDB and WordNet are two most useful resources for both SNLI and SciTail.

\section{In-Depth Analysis: D+R}
\begin{figure*}[ht]
\centering
\subfloat[\small{D+R (5\%) }]{\includegraphics[trim={0mm 0mm 0mm 0mm},clip,width=0.49\linewidth]{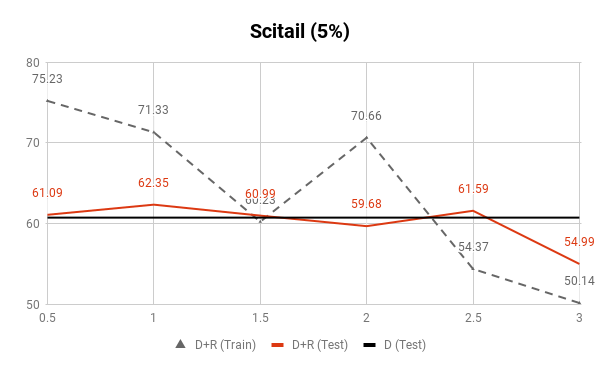}}
\subfloat[\small{D+R (10\%) }]{\includegraphics[trim={0mm 0mm 0mm 0mm},clip,width=0.49\linewidth]{figs/{DR_10_scitail}.png}}
\\
\subfloat[\small{D+R (25\%) }]{\includegraphics[trim={0mm 0mm 0mm 0mm},clip,width=0.49\linewidth]{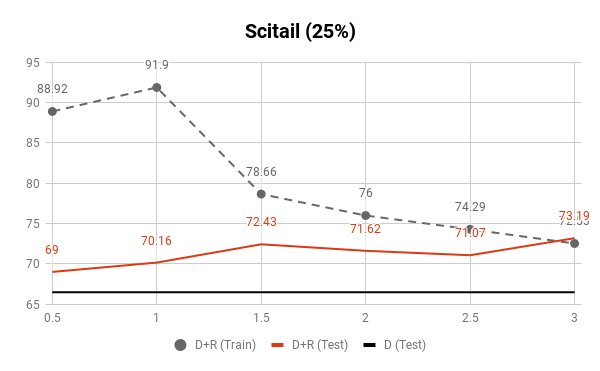}}
\subfloat[\small{D+R (50\%) }]{\includegraphics[trim={0mm 0mm 0mm 0mm},clip,width=0.49\linewidth]{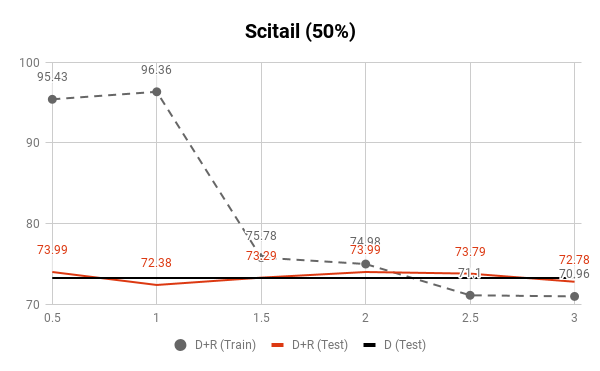}}
\\
\subfloat[\small{D+R (75\%) }]{\includegraphics[trim={0mm 0mm 0mm 0mm},clip,width=0.49\linewidth]{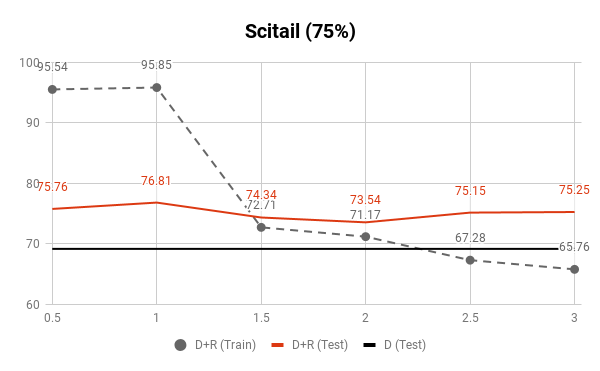}}
\subfloat[\small{D+R (100\%) }]{\includegraphics[trim={0mm 0mm 0mm 0mm},clip,width=0.49\linewidth]{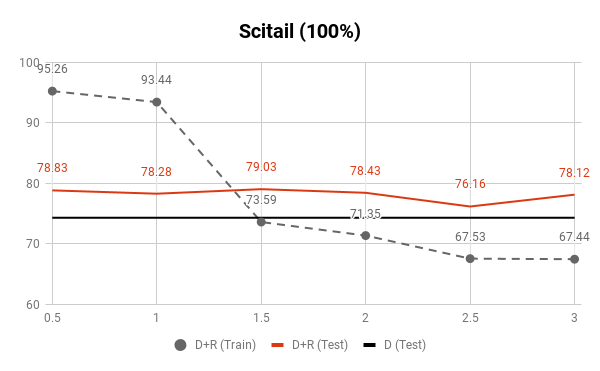}}
\caption{\disc+\genr with different ratio for SciTail. }  \label{fig:DR_scitail}
\end{figure*}

\begin{figure*}[ht]
\centering
\subfloat[\small{D+R (1\%) }]{\includegraphics[trim={0mm 0mm 0mm 0mm},clip,width=0.49\linewidth]{figs/{DR_1_snli}.png}}
% \subfloat[\small{D+R (2\%) }]{\includegraphics[trim={0mm 0mm 0mm 0mm},clip,width=0.24\linewidth]{figs/{DR_2_snli}.png}}
% \subfloat[\small{D+R (3\%) }]{\includegraphics[trim={0mm 0mm 0mm 0mm},clip,width=0.24\linewidth]{figs/{DR_3_snli}.png}}
\subfloat[\small{D+R (5\%) }]{\includegraphics[trim={0mm 0mm 0mm 0mm},clip,width=0.49\linewidth]{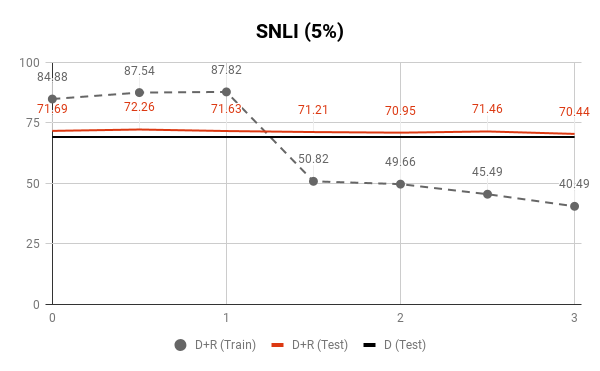}}
\\
\subfloat[\small{D+R (9\%) }]{\includegraphics[trim={0mm 0mm 0mm 0mm},clip,width=0.49\linewidth]{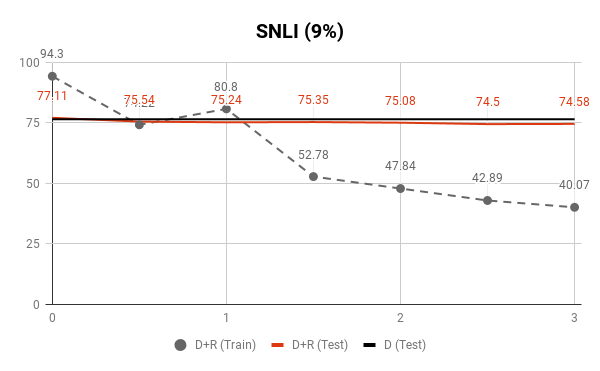}}
\subfloat[\small{D+R (25\%) }]{\includegraphics[trim={0mm 0mm 0mm 0mm},clip,width=0.49\linewidth]{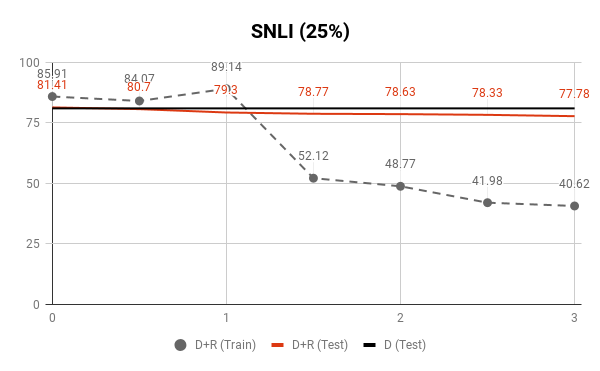}}
\\
\subfloat[\small{D+R (50\%) }]{\includegraphics[trim={0mm 0mm 0mm 0mm},clip,width=0.49\linewidth]{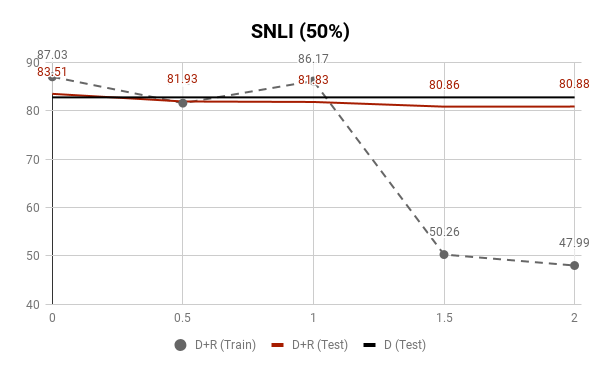}}
\subfloat[\small{D+R (100\%) }]{\includegraphics[trim={0mm 0mm 0mm 0mm},clip,width=0.49\linewidth]{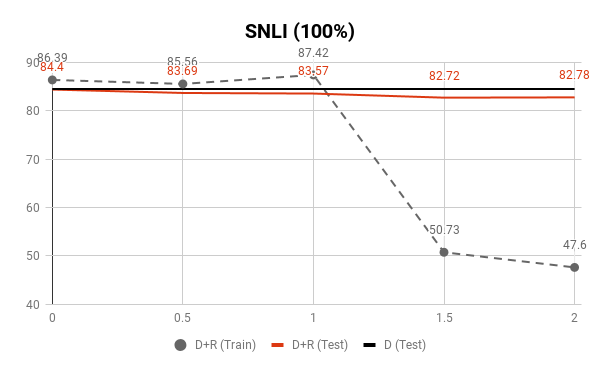}}
\caption{\disc+\genr with different ratio for SNLI. }  \label{fig:DR_snli}
\end{figure*}

Table~\ref{fig:DR_scitail} and Table~\ref{fig:DR_snli} show more in-depth analysis with different sub-sampling ratio on SNLI and SciTail.
The dotted line is training accuracy, and the solid red (\disc+\genr) and sold black (\disc) shows testing accuracies. 

\end{appendix}

\end{document}